\documentclass[wcp]{jmlr}

\usepackage{longtable}% for long tables

\usepackage{booktabs}
\usepackage[]{algorithm2e}
\usepackage{graphicx}
\graphicspath{ {./pdfs/} }
\usepackage{amsmath}
\usepackage{listings}
\usepackage{algorithm}
\usepackage{algorithmic}
\usepackage{xcolor}
\usepackage{bm}

\newcommand{\double}[1]{\mathbb{#1}}
\newcommand{\given}{\,|\,}

\definecolor{keywords}{RGB}{255,0,90}
\definecolor{comments}{RGB}{0,113,0}
\definecolor{red}{RGB}{160,0,0}
\definecolor{green}{RGB}{0,150,0}

\def\ie{\textit{i.e.}~}
\def\eg{\textit{e.g.}~}

\makeatletter
\let\Ginclude@graphics\@org@Ginclude@graphics 
\makeatother

\jmlrvolume{157}
\jmlryear{2021}
\jmlrworkshop{ACML 2021}

\title[Contrastive Neural Processes]{Contrastive Neural Processes for Self-Supervised Learning}

\author{\Name{Konstantinos Kallidromitis} \Email{k.kallidromitis@us.panasonic.com}\\
 \Name{Denis Gudovskiy} \Email{denis.gudovskiy@us.panasonic.com}\\
\addr Panasonic AI Lab, USA
 \AND
 \Name{Kazuki Kozuka} \Email{kozuka.kazuki@jp.panasonic.com}\\
 \addr Panasonic Technology Division, Japan
 \AND
 \Name{Iku Ohama} \Email{iku.ohama@us.panasonic.com}\\
 \addr Panasonic AI Lab, USA
 \AND
 \Name{Luca Rigazio} \Email{luca@aiolilabs.com}\\
 \addr AIoli Labs, USA}

\begin{document}

\maketitle

\begin{abstract}
Recent contrastive methods show significant improvement in self-supervised learning in several domains. In particular, contrastive methods are most effective where data augmentation can be easily constructed \eg in computer vision. However, they are less successful in domains without established data transformations such as time series data. In this paper, we propose a novel self-supervised learning framework that combines contrastive learning with neural processes. It relies on recent advances in neural processes to perform time series forecasting. This allows to generate augmented versions of data by employing a set of various sampling functions and, hence, avoid manually designed augmentations. We extend conventional neural processes and propose a new contrastive loss to learn times series representations in a self-supervised setup. Therefore, unlike previous self-supervised methods, our augmentation pipeline is task-agnostic, enabling our method to perform well across various applications. In particular, a ResNet with a linear classifier trained using our approach is able to outperform state-of-the-art techniques across industrial, medical and audio datasets improving accuracy over 10\% in ECG periodic data. We further demonstrate that our self-supervised representations are more efficient in the latent space, improving multiple clustering indexes and that fine-tuning our method on 10\% of labels achieves results competitive to fully-supervised learning.
\end{abstract}

\begin{figure}[ht]
\begin{center}
\includegraphics[width=1\textwidth]{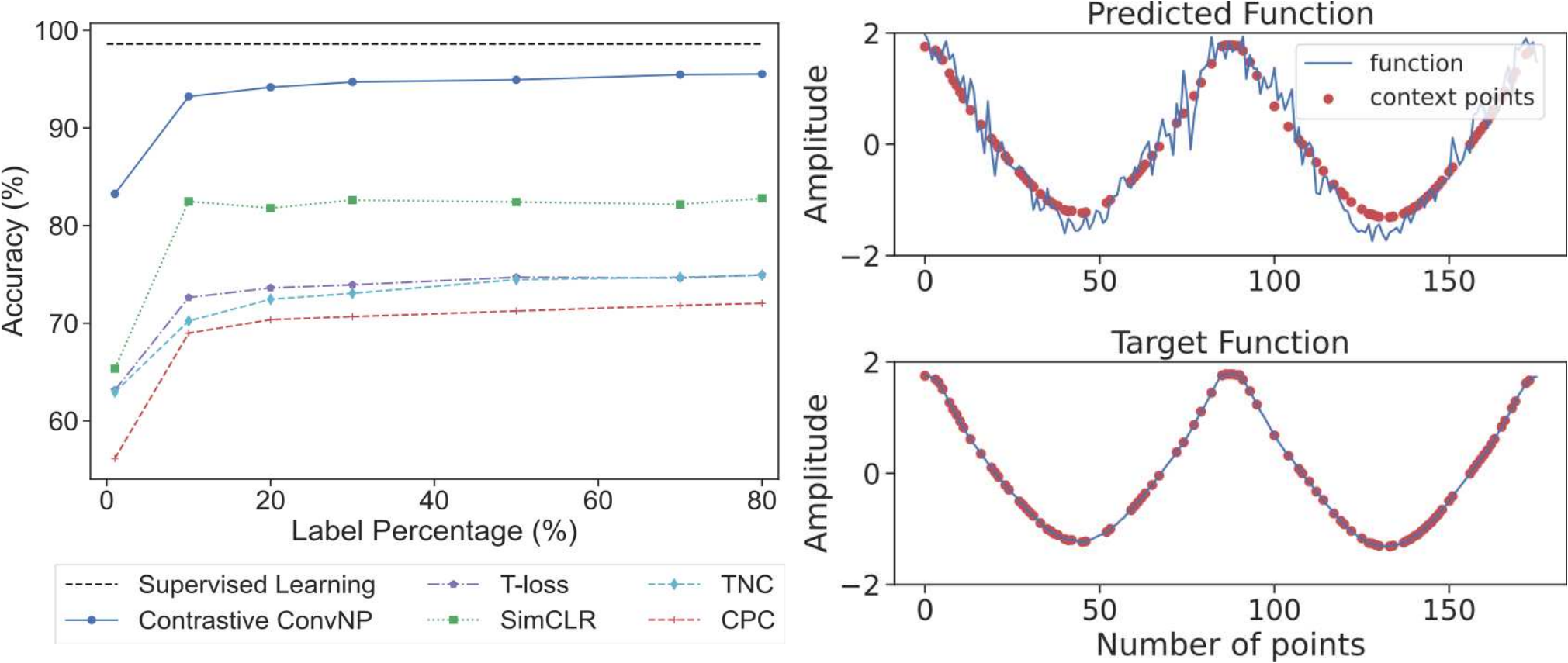}
\caption{(a) Downstream classification accuracy of ContrNP (Contrastive Neural Processes) compared with self-supervised and supervised learning baselines for different label percentages of the Atrial Fibrillation dataset (AFDB) using a linear classifier. Our method is shown to outperform other unsupervised approaches and benefit more at low label percentages. (b) Contrastive Neural Process predicting a sinusoidal function with only 40 context samples, showing that our loss function is able to predict not only contrastive but also forecasting elements.}
\label{fig:1}
\end{center}
\end{figure}

%-------------------------------------------------------------------------
\section{Introduction}
\label{sec:intro}
Self-supervised learning enables the learning of effective task-agnostic representations that generalize to a wide range of downstream applications without labels~\citep{simclr}. In particular, current self-supervised methods provide results competitive to supervised learning while using a fraction of labels for finetuning in the field of computer vision~\citep{byol, bigself}. This is achieved using a contrastive loss function from earlier works in contrastive learning~\citep{nce, similarity} and strong augmentation pipelines~\citep{amdim, moco} which improve regularization in low-label settings~\citep{howuseful}. While it is relatively easy to construct such augmentation pipelines in the image data domain using geometric transformations~\citep{swav, singleimage}, it is difficult to design augmentations in other domains \eg for time series data.

Time series models have many practical applications in manufacturing~\citep{imperfect, bearing}, financial forecasting~\citep{financial} and healthcare~\citep{clocs}, where data tends to be multivariate and highly imbalanced. Also, such models usually predict long data sequences using unlabeled or sparsely labeled datasets~\citep{ts_complex}. Recent self-supervised methods for time series data~\citep{tloss, cpc, tnc} learn representations using forecasting or distance-based metrics in order to employ conventional contrastive loss functions. Similarly, newer methods with established augmentation pipelines in the field of audio processing, attempt to use various signal transformations in the contrastive loss~\citep{audiosimclr, audiospk}. However, these advances cannot easily translate to every time series model, as each application benefits from a different set of augmentations. To address this issue, we propose an approach that relies on contrastive neural processes.

Neural Processes (NPs) are able to generate different agnostic representations by sampling a function multiple times. NPs first create separate embeddings for each observation and then aggregate them to a latent variable~\citep{np}. This general representation is then used to predict unobserved points from a target set, in order to model a distribution over regression functions~\citep{cnp, anp}. Our solution leverages the regression of the NPs as a supervised signal for unsupervised learning. More specifically, we can generate multiple observations of the same data point by applying different sampling functions. Each set of observations can be considered as different view of the same data point similar to an augmentation~\citep{fcrl}. This enables the use of any contrastive self-supervised loss to train our model.

Our method, Contrastive Neural Processes (ContrNP), employs the forecasting element of neural processes as a signal to extract better representations. We train the neural processes with a modified loss function that includes a contrastive term and a regression term allowing our method to distinguish the representations more efficiently in the latent space. ContrNP can be implemented with any type of data, but because of the lack of established augmentations and range of applications, time series provide a good benchmark. It follows methods like Contrastive Predictive Coding (CPC)~\citep{cpc}, which use an auto-regressive model to optimise the representation on an unsupervised prediction task, but in our case the autoregressive model is replaced by a neural process. One of the latest iterations in NPs is the convolutional conditional neural processes~\citep{convcnp}, which model translation equivariance and significantly boost forecasting performance on time series. This version of NPs is able to predict time series accurately which is combined with a contrastive learning loss to extract improved representations in the latent space. Our main findings and contributions can be summarized as follows:
\begin{itemize}
	\item We propose a new technique to perform self-supervised learning that uses neural processes as a forecasting model and contrastive learning to outperform current state-of-the-art models for non-stationary multivariate time series data.
	\item We further demonstrate that our method is able to achieve competitive results to supervised learning while requiring only a handful of labels and not requiring any specific augmentation techniques.
\end{itemize}

%-------------------------------------------------------------------------
\section{Related Work}
\label{sec:related}
\subsection{Self-Supervised Learning}
Self-supervised learning (SSL) uses an auxiliary task combined with an objective function to learn useful data representations~\citep{howuseful}. The contrastive predictive coding (CPC) approach~\citep{cpc} introduces an InfoNCE \textit{contrastive loss objective} and uses an autoregressive model to bring predictions closer to the actual value of the input in the latent space. Such SSL methods using contrastive loss quickly become state-of-the-art in computer vision~\citep{cpc2}. Their analysis shows that strong data augmentation pipelines are a crucial aspect to achieve optimal results~\citep{bigself}. For example, the most advanced methods combine multiple augmentation steps for each input~\citep{amdim} to further enhance data regularization.

\begin{figure}[ht]
	\begin{center}
		\includegraphics[width=1\textwidth]{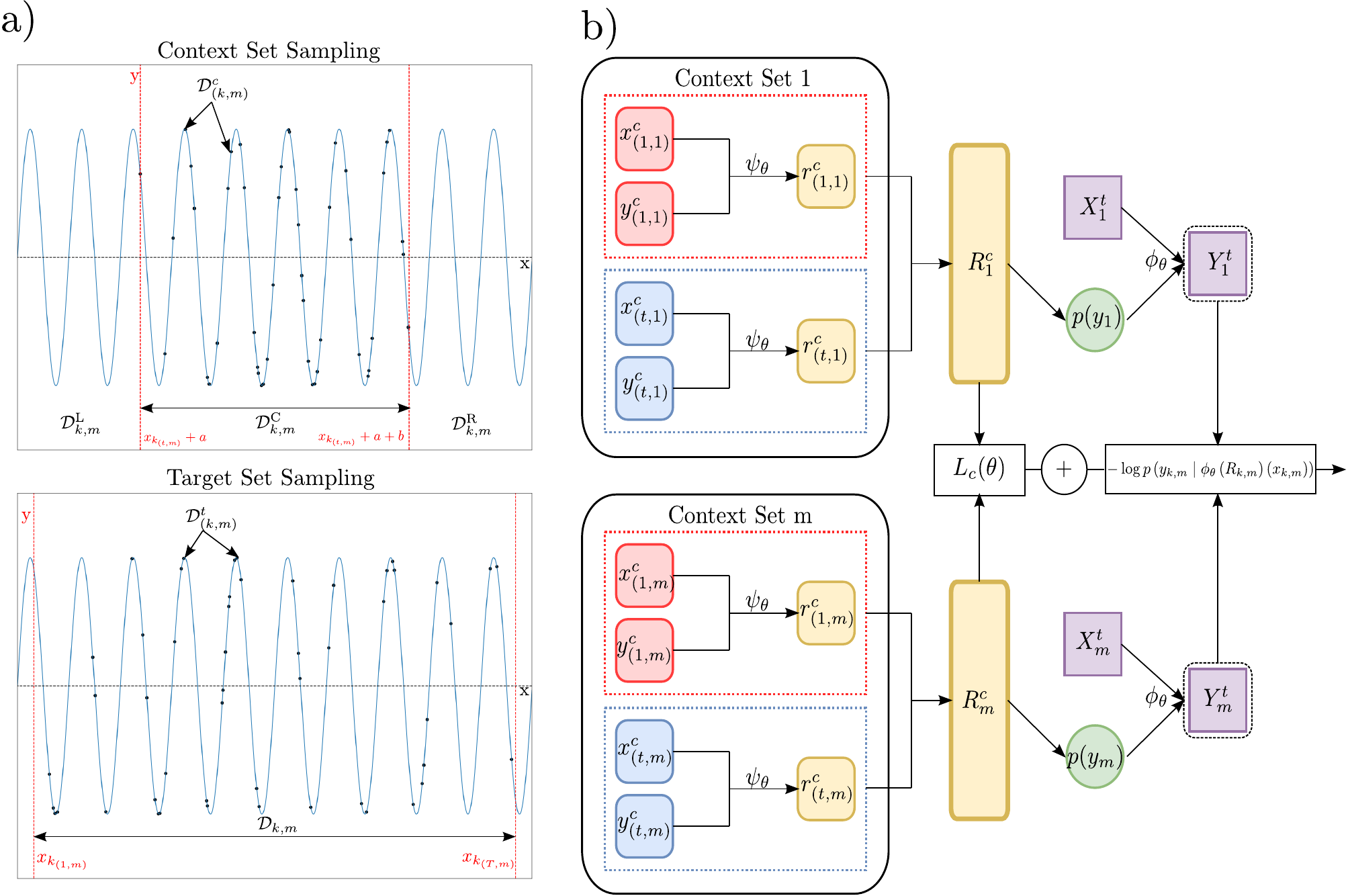}
		\caption{(a) Out of context sampling for a time series $\mathcal{D}$ showing the range $\mathcal{D}^C_{k,m}$ for the context set (up) and the range $\mathcal{D}_{k,m}$ for the target set (down), where $\mathcal{D}_{k,m} = \mathcal{D}_{k,m}^{\mathrm{L}} \cup \mathcal{D}_{k,m}^{\mathrm{C}} \cup \mathcal{D}_{k,m}^{\mathrm{R}}$. The reduced sampling range used in the context set improves generalization. (b) Our ContrNP approach for a segment $\mathcal{D}_k$ which has M context sets that pass through the neural process encoder and aggregate to an $R^c_m$ representation, and are finally used to compute the contrastive loss term $L_c$. NPs also decode the representations $R^c_1,R^c_m$ with the target set to perform forecasting and compute the negative log likelihood loss. Both metrics are combined to produce the final loss function used for training.}
		\label{fig:2}
	\end{center}
\end{figure}

Moreover, SimCLR~\citep{simclr} uses a loss where the input $x$ is passed through the pipeline that generates two augmented versions of the same input image ($x_i, x_j$). Then that positive pair passes through an encoder and a projection head, which are trained to maximize agreement between them. The rest of a mini-batch is used as negative examples and the distance to the pair is maximized. MoCo~\citep{moco} maintains a dictionary of negative samples that is used to increase disagreement and to be less dependent on mini-batch size. BYOL~\citep{byol} is able to utilize a loss function that does not directly employ contrastive learning, but uses batch normalization layers that operate as implicit negative samples.

\subsection{Representation Learning for Time Series}

Recent successes in time series methods depend on the property of continuity, meaning that in a sequence nearby values are more related compared to values further away. The T-loss model~\citep{tloss} uses a triplet loss, which maximizes the distance between negative examples that are chosen independently at random while minimizing the distance between the reference sequence and its subsets. This follows the success of word2vec, where similar sequences must also have close representations and randomly sampled sequences must differ~\citep{word2vec}. The temporal neighborhood coding (TNC)~\citep{tnc} maps the proximity of sequences in a “neighborhood” and transfers them to the encoding space while maximizing the distance of sequences far away. Following ideas from positive-unlabeled learning~\citep{pu}, specific weights are assigned for each sample according to its location. Nearby samples have a unit weight and are considered positive. Unlabeled examples have probability ($1-w$) of being negative and $w$ of being positive. Some time series methods~\citep{clocs} have also attempted to use some limited augmentations for medical applications~\citep{physionet} by extending the NCE loss to include multiple positive representations of the same patient.

In speech recognition, augmentations also play a key role in self-supervised learning. Significant improvements in speaker verification~\citep{voxceleb} were achieved by augmentation adversarial training~\citep{audiospk}, which uses augmentations to train a channel-invariant pretext task. Furthermore, conventional time masking and frequency shift~\citep{specaugment} transformations are widely used augmentations in audio classification~\citep{audioset}. \citet{audiosimclr} propose a contrastive learning framework, where the agreement objective is achieved between different formats of the same audio \ie between a raw waveform and the corresponding spectrogram, respectively. The goal of designing methods with generalized data transformations (GDT) for audio classification has also been explored by \cite{multimodal}. GDT is a framework that combines transformations with a contrastive loss to ensure an augmentation is invariant and the choice of sample is distinctive.

\subsection{Neural Processes}

Conditional Neural Processes (CNPs) was the first to introduce a method to combine neural networks and stochastic processes to approximate a distribution over functions~\citep{cnp}. This is achieved by encoding a context set of observations into an aggregated representation that is then decoded to foreca
st the function's target set. A recent approach called Neural Processes (NP) sample a random, latent variable that represents the global uncertainty to create a stochastic process, instead of using the representation directly~\citep{np}. Convolutional CNPs~\citep{convcnp} embed the time or spatial input (distance between two inputs) to a function space directly, instead of a finite vector space and hence provides translation equivariance. Finally, the concept of using functions for contrastive learning has been implemented by FCRL~\citep{fcrl}. The authors sample different sets of observations and use a contrastive loss function similar to NT-Xent~\citep{simclr}, that brings representations from the same function closer and from other functions further apart. The key difference with our approach is that FCRL do not combine their method with the existing regression techniques in NPs that are able to extract a better representation of the data, but rather only optimize an encoder model at train time.

%-------------------------------------------------------------------------
\section{Proposed Method}
\label{sec:method}

We introduce a framework for learning task agnostic representations using neural processes to bypass the augmentation step (Figure \ref{fig:2}). In our explanation we concentrate on a typical time series $\mathcal{D} = \{D_t\}_{t=1}^T =\{x_t, y_t\}_{t=1}^T$, where $x_t$ is time stamp and $y_t$ is the output that corresponds to $x_t$. Our objective is to learn a time-dependent representation $\bm{R}_t = \psi_{\theta}(D_t)$ that best represents the input-output relationships $\mathcal{F}: \mathcal{X} \rightarrow \mathcal{Y}$ in the latent space. Note that $\psi_{\theta}$ is an encoder with model parameters $\theta$. In order to represent the input time series $\mathcal{D}$ that changes over time, we split $\mathcal{D}$ into $K$ segments as $\mathcal{D} \rightarrow \{\mathcal{D}_1, \mathcal{D}_2, \ldots, \mathcal{D}_k,\ldots, \mathcal{D}_K\}$. Then, we assume input-output relationships within each segment $\mathcal{D}_k$ can be characterized by an individual function $f_k$. Furthermore, for each $\mathcal{D}_k$, we randomly sample the data into $M\in Z^+$ groups as $\{\mathcal{D}_{k,m}\}_{m=1}^M$. The above mentioned data arrangement means that two different subsets $\mathcal{D}_{k,m}$ and $\mathcal{D}_{k',m'}$ assuming $k=k'$ both represent a unique version of the same underlying function and for $k\neq k'$ different functions.

NPs sample two sets of observations, for the context $\mathcal{D}^c_{k,m}$ and target $\mathcal{D}^t_{k,m}$ sets. The context set is used to generate the representations and the target set to verify the predictions of the decoder. Our objective is to learn the underlying representation $R$ that best represents the input in the latent space. Our approach works with any framework from the Neural Process family and any type of input. NPs aim to learn a distribution over functions such as $f_k$ and approximate them using out of context sampling in the target set. For our specific implementation we use the convolutional conditional neural processes (ConvCNP)~\citep{convcnp} because of its advantages when extracting time series representations. Compared to its predecessors~\citep{cnp,np}, convolutional CNP is able to initially map the inputs $x_t$ individually from $y_t$ to a functional representation that corresponds to the difference between two inputs, which guarantees translational equivariance~\citep{convcnp}. Then the signal is discretized and passed into a CNN that is used to extract a latent space representation $R$. The motivation behind using a convolutional neural process is twofold. Firstly, time series tend to have a periodicity and secondly, we are predicting samples that are outside of the context range which makes translational equivariance a useful property that allows robust predictions outside of the normal range.

This is achieved with out of context sampling (Figure \ref{fig:2}). NP sample a set of observations for the context $\mathcal{D}^c$ and target sets $\mathcal{D}^t$. The context set is used to generate the representations and the target set to verify the predictions of the decoder. We make the logical assumption similar to other time series approaches~\citep{tloss,tnc} that in the input $\mathcal{D}$ the smaller the distance (in time) is between two points $D_t$ and $D_{t'}$, the more related they are. This means that for a subset $\mathcal{D}_{k,m} = \{x_i, y_i\}_{i=1}^{N_{k,m}}$ that has a total of $N_{k,m}$ samples, we only use a limited range of samples for the context set. More specifically, first, we split $\mathcal{D}_{k,m}$ in to three parts as $\mathcal{D}_{k,m}^{\mathrm{L}} = \left\{\{x_{i'}, y_{i'}\} \in \mathcal{D}_{k,m} ~|~ x_{i'} \leq a\right\}$, $\mathcal{D}_{k,m}^{\mathrm{C}} = \left\{\{x_{i'}, y_{i'}\} \in \mathcal{D}_{k,m} ~|~ a < x_{i'} < b\right\}$, and $\mathcal{D}_{k,m}^{\mathrm{R}} = \left\{\{x_{i'}, y_{i'}\} \in \mathcal{D}_{k,m} ~|~ b \leq x_{i'}\right\}$, where $a$ and $b$ are thresholds. Then only $\mathcal{D}_{k,m}^{\mathrm{C}}$ is used for context set. Thus, the thresholds $a,b$ are typically chosen so that context set include intended number of examples. While the entire range $\mathcal{D}_{k,m} = \mathcal{D}_{k,m}^{\mathrm{L}} \cup \mathcal{D}_{k,m}^{\mathrm{C}} \cup \mathcal{D}_{k,m}^{\mathrm{R}}$ for the target set. This allows the context set observations $\mathcal{D}_{k,m}^{\mathrm{c}}$ to represent the function $f_k$, but because we exclude a large part of the range $\mathcal{D}_{k,m}^{\mathrm{L}},\mathcal{D}_{k,m}^{\mathrm{R}}$ we are predicting on, there is higher regularization which is equivalent to using a stronger (larger) augmentation. This follows with conventional theory that states stronger data augmentations are able to further improve generalization and the performance of a self-supervised learning framework~\citep{simclr}. Our method thus allows data agnostic augmentations by taking advantage of sampling.

After extracting the main representation $R_{k,m}$ of the function $f_k$ by passing the context points $\mathcal{D}_{k,m}^{\mathrm{c}}$ in the encoder of the neural process, we also repeat the step for the second version $\mathcal{D}_{k,m'}^{\mathrm{c}}$ of the function $f_k$ and extract $R_{k,m'}$ where $m\neq m'$. The two representations $R_{k,m}$ and $R_{k,m'}$ can be used for the contrastive term of the loss function as the anchor and positive example respectively. Similarly, all the other representations in the batch $\bm{R}_{k',\cdot}$ where $k\neq k'$ are the negative examples. Equation~\ref{eq:1} is the alternate version of the NT-Xent loss used in the SimCLR paper~\citep{simclr}, but using sampling functions instead of augmentations \citep{fcrl}. For an encoder $\psi_{\theta}$, the anchor representation $R_{k,m} = \psi_{\theta}\left( \mathcal{D}_{k,m} \right)$. The equation minimizes the distance between the anchor $R_{k,m}$ and a positive example $R_{k,m'}$, and maximizes the distance of the anchor $R_{k,m}$ and all negative examples $\bm{R}_{k',\cdot}$ in the batch. More specifically, we define our contrastive loss as: 

\begin{equation}\label{eq:1}
\mathcal{L}_{\mathrm{C}}(\bm{\theta},\mathcal{D}_{1:K}) = \sum_{k=1}^K\sum_{m=1}^M\sum_{m'=1}^M\double{I}_{[m\neq m']}
\log\left[{\frac{\mbox{Sim}(R_{k,m}, R_{k, m'})/\tau}
{\sum\limits_{k'=1}^K \sum\limits_{m''=1}^M\double{I}_{[k'\neq k]}
\mbox{Sim}(R_{k,m}, R_{k', m''})/\tau}}\right]
\end{equation}

where $\mbox{Sim}(\cdot, \cdot)$ is cosine similarity, $\tau$ is temperature parameter~\citep{simclr} and $\double{I}_{[\cdot]}$ is indicator function that returns $1$ if given condition holds and 0 otherwise.

After extracting the context representation $R_{k,m}$, we use the neural process decoder $\phi_\theta (\cdot)$ to predict the mean and standard deviation of $\bm{y}_{k,m}$ given $\bm{x}_{k,m}$. We compute the loss using Equation~\ref{eq:2}, where we combine the negative log likelihood loss together with the contrastive term $\mathcal{L}_{\mathrm{C}} (\bm{\theta}, \mathcal{D}_{1:K})$ computed from the representations. The initial part of Equation~\ref{eq:2} is weighted by $\lambda$ which determines the ratio between the first term which is trained to perform interpolation within the function and the second which is used to distinguish between the different functions. The output of the decoder is compared to the theoretical value of $\bm{y}_{k,m}$ to extract better representations.

\begin{equation}\label{eq:2}
\mathcal{L} (\theta) =- \lambda\sum_{k=1}^K\sum_{m=1}^M\log P \left( \bm{y}_{k,m} \given \phi_{\theta} (R_{k,m}), \bm{x}_{k,m}\right)+\underbrace{\mathcal{L}_{\mathrm{C}} (\bm{\theta}, \mathcal{D}_{1:K})}_{\text{contrastive term}}
\end{equation}

In summary, for our implementation we use the off-grid ConvCNP algorithm for signal interpolation. During the computation of the loss function we add a contrastive term that brings the representations from the same function closer while distancing the ones from other functions. In addition to the contrastive learning, our method also takes advantage of the advances in neural processes to learn the within characteristics of the signal and create efficient representations in the latent space. In order to compare our model in downstream tasks such as classification against supervised learning, we run it at test time and extract representations from the encoder which we aggregate into a single one $R_k = \frac{1}{M}\sum_{m=1}^M R_{k,m}$ for each function $f_k$. The representations are then fed into a ssingle layer classifier that is trained independently.

%-------------------------------------------------------------------------
\section{Experiments}
\label{sec:eval}
To illustrate the advantages of contrastive neural processes we consider two different experimental setups involving noisy and complex time series from medical and industrial domains to examine their performance on real-world datasets. The experiments cover a variety of applications and metrics that show the adaptability of our agnostic approach. The first experiment is a downstream classification task with a linear classifier which evaluates the model's performance on the AFDB, IMS Bearings and Urban8K datasets. For the second experiment we assess the representations directly using Silhouette score and Davies–Bouldin index (DBI) similar to~\citep{tnc}. All models are implemented in PyTorch (version 1:3:1) and trained on Tesla P100-16GB GPUs.The code corresponding to these experiments can be found in the supplementary material. For each implementation we trained in an unsupervised way with the entire train set. For evaluation, we used 80\% of the train labels to finetune a decoder and the remaining labels were are used in the validation set to select the results with the best validation accuracy. Below we describe each of our experiments in more detail. 

\textbf{MIT-BIH Atrial Fibrillation} (AFDB) \footnote{\url{https://physionet.org/content/afdb/1.0.0/}} is a collection of 25 electrocardiogram (ECG) recordings where each recording has a duration of approximately 10 hours~\citep{moody}. The recordings have two ECG signal types sampled at 250 Hz. The dataset includes 4 classes: (1) Atrial fibrillation, (2) Atrial flutter, (3) AV junctional rhythm and (4) all other rhythms. The dataset was chosen due to its long duration and changing properties as time progresses (alternating classes). Moreover, it is highly imbalanced, for instance, class 3 only appears in 0.1\% of the data-points~\citep{tnc}.

\textbf{IMS Bearing} \footnote{\url{https://ti.arc.nasa.gov/tech/dash/groups/pcoe/prognostic-data-repository/}} is a dataset that was collected from a run-to-failure experiment on 4 bearings rotating with 2000 rpm on a loaded shaft 6000 lbs~\citep{bearingdata}. A similar setup to~\citep{imperfect} was followed, using only the third bearing in our experiments. The data was split into 5 classes where each indicated the health state (eg. early, normal, imminent failure) of the bearing. The experiment included two high precision accelerometers (x-axis, y-axis). The dataset was selected to examine the performance of the methods across a long and noisy industrial time series.

\textbf{Urban8K} \footnote{\url{https://https://urbansounddataset.weebly.com/urbansound8k}} is a dataset that contains 8,732 urban sound clips of varying size ($<=4s$)~\citep{urban8k}. It consists of 10 classes such as children playing, car horn, dog bark and street music. The files were taken from field recordings and were presorted into 10 folds. Each evaluation metric was calculated by averaging the 10-fold cross-validation performed using the preexisting folds.

\subsection{Downstream Classification}
For the first experiment we compared the accuracy and AUCPRC using a CNN encoder architecture.  Our method significantly outperformed existing methodologies by more than 10\% in the AFDB dataset, containing periodic time signals. One of the strengths of our approach, is the ability to learn using a forecasting objective similar to Convolutional CNPs, which perform well on periodic time signals due to their translation equivariance~\citep{convcnp}. This is further demonstrated in Appendix \ref{apd:a1} where we display the regression output of our model. Similarly, for the IMS and Urban8K datasets, our model was able to outperform other unsupervised baselines. Our experiments found that SimCLR achieved the lowest accuracy, 41.5\% as seen in Table \ref{tab:accuracy}, in the IMS dataset, due to its augmentation pipeline. The standard augmentations of Time-warp and Frequency Shift proved to be insufficient and the score can be improved by experimenting with a different augmentations (Appendix \ref{apd:a2}). Even though contrastive neural processes have a similar loss function to SimCLR they achieved the highest performance without any need for augmentations.

\begin{table}[ht]
\caption{Classification test accuracy and AUC for AFDB, IMS Bearing and Urban8K datasets, each one indicates the average of 5, 5 and 10 runs respectively $\mu_{\pm\sigma}$\%. Self-supervised experiments are trained on the entire train set and then for the downstream classification task 80\% of the labels are used for fine-tuning.}
%69.4\tiny$\pm$0.2 & 84.4\tiny$\pm$0.1
\label{tab:accuracy}
\centering
\begin{tabular}{lcccccc}
    & \multicolumn{2}{c}{AFDB} & \multicolumn{2}{c}{IMS Bearing}& \multicolumn{2}{c}{Urban8K} \\
    \toprule
    Method    & Accuracy  & AUPRC  & Accuracy  & AUPRC & Accuracy  & AUPRC \\
    \midrule
    CPC & 71.6\tiny$\pm$0.2 & 62.6\tiny$\pm$0.4 & 72.4\tiny$\pm$0.1 & 84.4\tiny$\pm$0.0 & 83.3\tiny$\pm$0.1 & 94.5\tiny$\pm$0.1 \\
    Tloss & 74.8\tiny$\pm$0.1 & 59.8\tiny$\pm$0.5 & 73.2\tiny$\pm$0.2 & 87.6\tiny$\pm$0.1 & 81.5\tiny$\pm$0.4 & 93.8\tiny$\pm$0.3 \\
    TNC & 74.5\tiny$\pm$0.4 & 56.3\tiny$\pm$0.4 & 70.3\tiny$\pm$0.3 & 86.3\tiny$\pm$0.1 & 80.7\tiny$\pm$0.1 & 93.9\tiny$\pm$0.4 \\
    SimCLR & 82.3\tiny$\pm$0.1 & 71.5\tiny$\pm$0.1 & 41.5\tiny$\pm$0.2 & 70.7\tiny$\pm$0.1 & 82.8\tiny$\pm$0.2 & 94.1\tiny$\pm$0.1\\
    ContrNP (ours) & \textbf{94.2}\tiny$\pm$0.4 & \textbf{89.1}\tiny$\pm$0.9 & \textbf{73.6}\tiny$\pm$0.1 & \textbf{89.3}\tiny$\pm$0.2 & \textbf{84.2}\tiny$\pm$0.5 & \textbf{95.4}\tiny$\pm$0.4\\
    \midrule
    Fully supervised & 98.4\tiny$\pm$0.0 & 81.6\tiny$\pm$0.2 & 86.3\tiny$\pm$0.0 & 94.8\tiny$\pm$0.1 & 99.9\tiny$\pm$0.0 & 99.9\tiny$\pm$0.0 \\
    \bottomrule
\end{tabular}
\end{table}

%-------------------------------------------------------------------------
\subsection{Clustering}
For the second experiment we examined the clusterability of multivariate time series datasets using self-supervised baselines and supervised learning. The metrics we employed evaluated the quality of the representations extracted from unsupervised models in a task agnostic way without the use of a downstream task. 

\begin{figure}[ht]
\begin{center}
\includegraphics[width=1\textwidth]{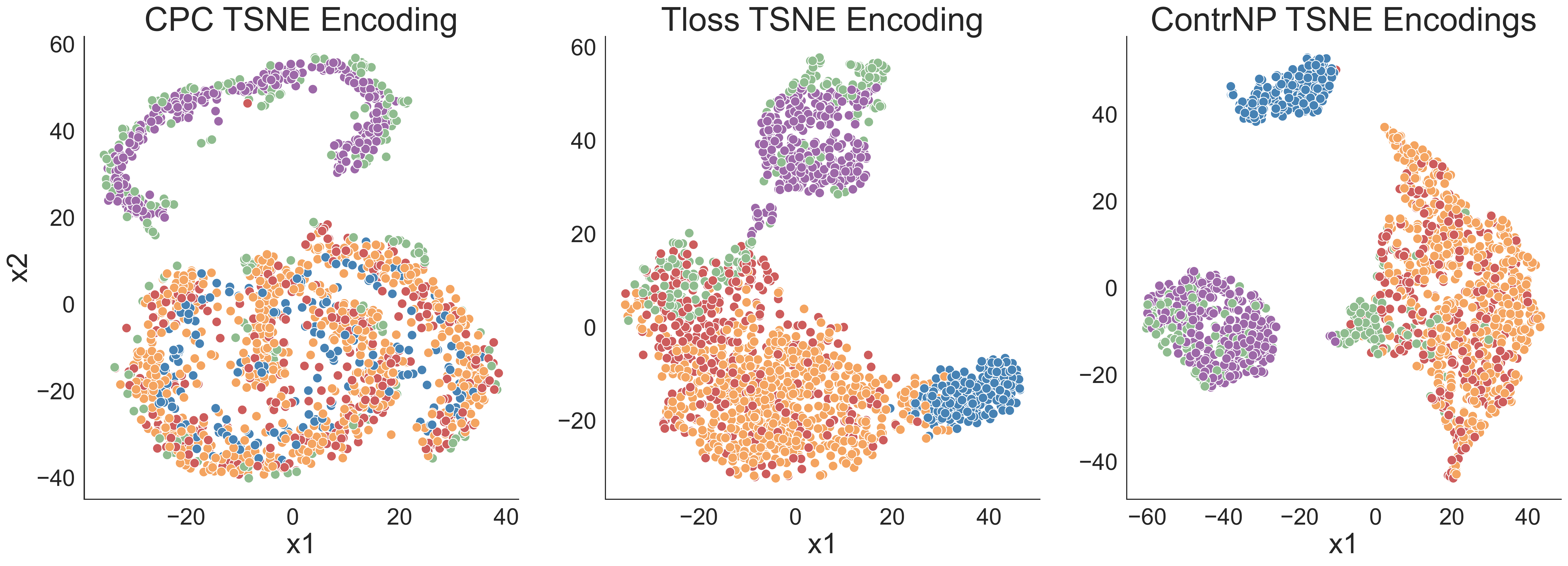}
\caption{TSNE 2D Encoding Representations extracted for the IMS dataset using unsupervised training with the CPC (left), Tloss(middle) and ContrNP (right) methods.}
\label{fig:3}
\end{center}
\end{figure}

We employed the Silhouette score that measures the proximity of points within the same cluster and the distance to points in neighbouring clusters. The values range between $[-1, +1]$, with higher values implying better clustering. The Davies–Bouldin index (DBI) is the average ratio of within cluster similarity over the similarity of its most similar cluster~\citep{tnc}. A smaller value indicates a low intra-cluster scatter and a high inter-cluster difference and signifies better separation of the representations. Figure \ref{fig:3} shows the TSNE encodings of CPC, Tloss and our approach respectively. The CPC encoding was able to distinguish some basic characteristics of the classes but failed to separate them well. It can be seen that even though our approach outperformed Tloss by a small margin in terms of accuracy, its encoding res presentations were better separated. This also aligns with the results from Table \ref{tab:clusters} where ContrNP outperformed Tloss in both clustering metrics. This is likely due to our dual objective function that learns to separate representations of different classes and also the within class characteristics.

\begin{table}[ht]
\caption{Silhouette score (Sil) and Davies–Bouldin index (DBI) were used to measure the cluster performance on the test set encodings of AFDB, IMS Bearing and Urban8K datasets with a 80\% train and 20\% test split, measured for 5, 5 and 10 total runs respectively $\mu_{\pm\sigma}$. A high silhouette score indicates good representations. It shows a small distance between the points in the same clusters and a high distance to other clusters. DBI measures the within similarity of cluster over the similarity with the closest one and improves as value decreases.}

\label{tab:clusters}
\centering
\begin{tabular}{lcccccc}
    & \multicolumn{2}{c}{AFDB} & \multicolumn{2}{c}{IMS Bearing}& \multicolumn{2}{c}{Urban8K} \\
    \toprule
    Method          & Sil $\uparrow$  & DBI $\downarrow$ & Sil $\uparrow$  & DBI $\downarrow$ & Sil $\uparrow$  & DBI $\downarrow$\\
    \midrule
    CPC & 0.22\tiny$\pm$0.02 & 1.74\tiny$\pm$0.10 & 0.12\tiny$\pm$0.01 & 2.20\tiny$\pm$0.05 & 0.24\tiny$\pm$0.04 & 1.64\tiny$\pm$0.25 \\
    Tloss & 0.14\tiny$\pm$0.03 & 2.04\tiny$\pm$0.07 & 0.17\tiny$\pm$0.01 & 1.79\tiny$\pm$0.15 & 0.26\tiny$\pm$0.03 & 1.30\tiny$\pm$0.08 \\
    TNC & 0.24\tiny$\pm$0.03 & 1.44\tiny$\pm$0.22 & 0.31\tiny$\pm$0.07 & 0.94\tiny$\pm$0.16 &0.36\tiny$\pm$0.05 & \textbf{0.72}\tiny$\pm$0.08 \\
    SimCLR & 0.34\tiny$\pm$0.02 & 1.49\tiny$\pm$0.20 & 0.24\tiny$\pm$0.04 & 1.47\tiny$\pm$0.20 & 0.35\tiny$\pm$0.05 & 1.13\tiny$\pm$0.12 \\
    ContrNP (ours) & \textbf{0.36}\tiny$\pm$0.07 & \textbf{1.35}\tiny$\pm$0.16 & \textbf{0.38}\tiny$\pm$0.06 & \textbf{0.91}\tiny$\pm$0.14 & \textbf{0.42}\tiny$\pm$0.06 & 0.89\tiny$\pm$0.15\\
    \midrule
    Fully supervised & 0.43\tiny$\pm$0.04 & 0.83\tiny$\pm$0.1 & 0.47\tiny$\pm$0.02 & 0.77\tiny$\pm$0.04 & 0.49\tiny$\pm$0.02 & 0.80\tiny$\pm$0.05\\
    \bottomrule
\end{tabular}
\end{table}

%-------------------------------------------------------------------------
\section{Conclusion}
\label{sec:conclusion}
This paper presents a novel method for self-supervised learning that does not require augmentation engineering. Time series present an ideal area to test our approach because of its lack of established augmentations and variety of different applications. Our objective function uses the forecasting objective of the neural process as a supervision signal and combines it with a contrastive loss that minimizes the distance between two representations of the same input and maximizes distance with the batch. We also propose out of context sampling to increase regularization in the model and improve the quality of augmentations. Finally, we show that our method is able to produce state of the art results across a wide range of difficult, real-world applications and reach the performance of supervised learning in some with only a fraction of the labels.

\bibliography{main}

\begin{thebibliography}{36}
\providecommand{\natexlab}[1]{#1}
\providecommand{\url}[1]{\texttt{#1}}
\expandafter\ifx\csname urlstyle\endcsname\relax
  \providecommand{\doi}[1]{doi: #1}\else
  \providecommand{\doi}{doi: \begingroup \urlstyle{rm}\Url}\fi

\bibitem[Asano et~al.(2020)Asano, Rupprecht, and Vedaldi]{singleimage}
Yuki~M. Asano, Christian Rupprecht, and Andrea Vedaldi.
\newblock A critical analysis of self-supervision, or what we can learn from a
  single image.
\newblock \emph{arXiv:1904.13132}, 2020.

\bibitem[Bachman et~al.(2019)Bachman, Hjelm, and Buchwalter]{amdim}
Philip Bachman, R~Devon Hjelm, and William Buchwalter.
\newblock Learning representations by maximizing mutual information across
  views.
\newblock In \emph{Advances in Neural Information Processing Systems}, 2019.

\bibitem[Caron et~al.(2020)Caron, Misra, Mairal, Goyal, Bojanowski, and
  Joulin]{swav}
Mathilde Caron, Ishan Misra, Julien Mairal, Priya Goyal, Piotr Bojanowski, and
  Armand Joulin.
\newblock Unsupervised learning of visual features by contrasting cluster
  assignments.
\newblock In \emph{Advances in Neural Information Processing Systems}, 2020.

\bibitem[Chen et~al.(2020{\natexlab{a}})Chen, Kornblith, Norouzi, and
  Hinton]{simclr}
Ting Chen, Simon Kornblith, Mohammad Norouzi, and Geoffrey Hinton.
\newblock A simple framework for contrastive learning of visual
  representations.
\newblock \emph{arXiv:2002.05709}, 2020{\natexlab{a}}.

\bibitem[Chen et~al.(2020{\natexlab{b}})Chen, Kornblith, Swersky, Norouzi, and
  Hinton]{bigself}
Ting Chen, Simon Kornblith, Kevin Swersky, Mohammad Norouzi, and Geoffrey~E
  Hinton.
\newblock Big self-supervised models are strong semi-supervised learners.
\newblock In \emph{Advances in Neural Information Processing Systems},
  2020{\natexlab{b}}.

\bibitem[{Chopra} et~al.(2005){Chopra}, {Hadsell}, and {LeCun}]{similarity}
S.~{Chopra}, R.~{Hadsell}, and Y.~{LeCun}.
\newblock Learning a similarity metric discriminatively, with application to
  face verification.
\newblock In \emph{Proceedings of the IEEE/CVF Conference on Computer Vision
  and Pattern Recognition (CVPR)}, 2005.

\bibitem[Elkan and Noto(2008)]{pu}
Charles Elkan and Keith Noto.
\newblock Learning classifiers from only positive and unlabeled data.
\newblock In \emph{Proceedings of the 14th ACM SIGKDD International Conference
  on Knowledge Discovery and Data Mining}, 2008.

\bibitem[Franceschi et~al.(2019)Franceschi, Dieuleveut, and Jaggi]{tloss}
Jean-Yves Franceschi, Aymeric Dieuleveut, and Martin Jaggi.
\newblock Unsupervised scalable representation learning for multivariate time
  series.
\newblock In \emph{Advances in Neural Information Processing Systems}, 2019.

\bibitem[Garnelo et~al.(2018{\natexlab{a}})Garnelo, Rosenbaum, Maddison,
  Ramalho, Saxton, Shanahan, Teh, Rezende, and Eslami]{cnp}
Marta Garnelo, Dan Rosenbaum, Christopher Maddison, Tiago Ramalho, David
  Saxton, Murray Shanahan, Yee~Whye Teh, Danilo Rezende, and S.~M.~Ali Eslami.
\newblock Conditional neural processes.
\newblock In \emph{Proceedings of the International Conference on Machine
  Learning (ICML)}, 2018{\natexlab{a}}.

\bibitem[Garnelo et~al.(2018{\natexlab{b}})Garnelo, Schwarz, Rosenbaum, Viola,
  Rezende, Eslami, and Teh]{np}
Marta Garnelo, Jonathan Schwarz, Dan Rosenbaum, Fabio Viola, Danilo~J. Rezende,
  S.~M.~Ali Eslami, and Yee~Whye Teh.
\newblock Neural processes.
\newblock \emph{arXiv:1807.01622}, 2018{\natexlab{b}}.

\bibitem[Gemmeke et~al.(2017)Gemmeke, Ellis, Freedman, Jansen, Lawrence, Moore,
  Plakal, and Ritter]{audioset}
Jort~F. Gemmeke, Daniel P.~W. Ellis, Dylan Freedman, Aren Jansen, Wade
  Lawrence, R.~Channing Moore, Manoj Plakal, and Marvin Ritter.
\newblock Audio set: An ontology and human-labeled dataset for audio events.
\newblock In \emph{Proc. IEEE ICASSP 2017}, 2017.

\bibitem[Goldberg and Levy(2014)]{word2vec}
Yoav Goldberg and Omer Levy.
\newblock word2vec explained: deriving mikolov et al.'s negative-sampling
  word-embedding method.
\newblock \emph{arXiv:1402.3722}, 2014.

\bibitem[Goldberger et~al.(2000)Goldberger, Amaral, Glass, Hausdorff, Ivanov,
  Mark, Mietus, Moody, Peng, and Stanley]{physionet}
Ary~L. Goldberger, Luis A.~N. Amaral, Leon Glass, Jeffrey~M. Hausdorff,
  Plamen~Ch. Ivanov, Roger~G. Mark, Joseph~E. Mietus, George~B. Moody,
  Chung-Kang Peng, and H.~Eugene Stanley.
\newblock Physiobank, physiotoolkit, and physionet.
\newblock \emph{Circulation}, 2000.

\bibitem[Gondal et~al.(2021)Gondal, Joshi, Rahaman, Bauer, Wuthrich, and
  Sch{\"o}lkopf]{fcrl}
Muhammad~Waleed Gondal, Shruti Joshi, Nasim Rahaman, Stefan Bauer, Manuel
  Wuthrich, and Bernhard Sch{\"o}lkopf.
\newblock Function contrastive learning of transferable representations.
\newblock \emph{arXiv:2010.07093}, 2021.

\bibitem[Gordon et~al.(2020)Gordon, Bruinsma, Foong, Requeima, Dubois, and
  Turner]{convcnp}
Jonathan Gordon, Wessel~P. Bruinsma, Andrew Y.~K. Foong, James Requeima, Yann
  Dubois, and Richard~E. Turner.
\newblock Convolutional conditional neural processes.
\newblock In \emph{Proceedings of the International Conference on Learning
  Representations (ICLR)}, 2020.

\bibitem[Grabocka and Schmidt-Thieme(2014)]{ts_complex}
Josif Grabocka and Lars Schmidt-Thieme.
\newblock Invariant time-series factorization.
\newblock \emph{Data Mining and Knowledge Discovery}, 2014.

\bibitem[Grill et~al.(2020)Grill, Strub, Altch\'{e}, Tallec, Richemond,
  Buchatskaya, Doersch, Avila~Pires, Guo, Gheshlaghi~Azar, Piot, kavukcuoglu,
  Munos, and Valko]{byol}
Jean-Bastien Grill, Florian Strub, Florent Altch\'{e}, Corentin Tallec, Pierre
  Richemond, Elena Buchatskaya, Carl Doersch, Bernardo Avila~Pires, Zhaohan
  Guo, Mohammad Gheshlaghi~Azar, Bilal Piot, koray kavukcuoglu, Remi Munos, and
  Michal Valko.
\newblock Bootstrap your own latent - a new approach to self-supervised
  learning.
\newblock In \emph{Advances in Neural Information Processing Systems}, 2020.

\bibitem[Gutmann and Hyvärinen(2010)]{nce}
Michael Gutmann and Aapo Hyvärinen.
\newblock Noise-contrastive estimation: A new estimation principle for
  unnormalized statistical models.
\newblock In \emph{Proceedings of the International Conference on Artificial
  Intelligence and Statistics}, 2010.

\bibitem[He et~al.(2020)He, Fan, Wu, Xie, and Girshick]{moco}
Kaiming He, Haoqi Fan, Yuxin Wu, Saining Xie, and Ross Girshick.
\newblock Momentum contrast for unsupervised visual representation learning.
\newblock In \emph{Proceedings of the IEEE/CVF Conference on Computer Vision
  and Pattern Recognition (CVPR)}, 2020.

\bibitem[Henaff(2020)]{cpc2}
Olivier Henaff.
\newblock Data-efficient image recognition with contrastive predictive coding.
\newblock In \emph{Proceedings of the International Conference on Machine
  Learning (ICML)}, 2020.

\bibitem[Huh et~al.(2020)Huh, Heo, Kang, Watanabe, and Chung]{audiospk}
Jaesung Huh, Hee~Soo Heo, Jingu Kang, Shinji Watanabe, and Joon~Son Chung.
\newblock Augmentation adversarial training for self-supervised speaker
  recognition.
\newblock \emph{arXiv:2007.12085}, 2020.

\bibitem[{Kang} et~al.(2015){Kang}, {Kim}, {Wills}, and {Kim}]{bearing}
M.~{Kang}, J.~{Kim}, L.~M. {Wills}, and J.~{Kim}.
\newblock Time-varying and multiresolution envelope analysis and discriminative
  feature analysis for bearing fault diagnosis.
\newblock \emph{IEEE Transactions on Industrial Electronics}, 2015.

\bibitem[Kim et~al.(2019)Kim, Mnih, Schwarz, Garnelo, Eslami, Rosenbaum,
  Vinyals, and Teh]{anp}
Hyunjik Kim, Andriy Mnih, Jonathan Schwarz, Marta Garnelo, Ali Eslami, Dan
  Rosenbaum, Oriol Vinyals, and Yee~Whye Teh.
\newblock Attentive neural processes.
\newblock \emph{arXiv:1901.05761}, 2019.

\bibitem[Kim(2003)]{financial}
Kyoung-Jae Kim.
\newblock Financial time series forecasting using support vector machines.
\newblock \emph{Neurocomputing}, 2003.

\bibitem[Kiyasseh et~al.(2020)Kiyasseh, Zhu, and Clifton]{clocs}
Dani Kiyasseh, Tingting Zhu, and David~A. Clifton.
\newblock Clocs: Contrastive learning of cardiac signals across space, time,
  and patients.
\newblock \emph{arXiv:2005.13249}, 2020.

\bibitem[Moody(1983)]{moody}
George Moody.
\newblock A new method for detecting atrial fibrillation using rr intervals.
\newblock \emph{Computers in Cardiology}, 1983.

\bibitem[Nagrani et~al.(2017)Nagrani, Chung, and Zisserman]{voxceleb}
Arsha Nagrani, Joon~Son Chung, and Andrew Zisserman.
\newblock Voxceleb: A large-scale speaker identification dataset.
\newblock \emph{Interspeech 2017}, 2017.

\bibitem[Newell and Deng(2020)]{howuseful}
Alejandro Newell and Jia Deng.
\newblock How useful is self-supervised pretraining for visual tasks?
\newblock In \emph{Proceedings of the IEEE/CVF Conference on Computer Vision
  and Pattern Recognition (CVPR)}, 2020.

\bibitem[Park et~al.(2019)Park, Chan, Zhang, Chiu, Zoph, Cubuk, and
  Le]{specaugment}
Daniel~S. Park, William Chan, Yu~Zhang, Chung-Cheng Chiu, Barret Zoph, Ekin~D.
  Cubuk, and Quoc~V. Le.
\newblock Specaugment: A simple data augmentation method for automatic speech
  recognition.
\newblock \emph{Interspeech 2019}, 2019.

\bibitem[Patrick et~al.(2020)Patrick, Asano, Kuznetsova, Fong, Henriques,
  Zweig, and Vedaldi]{multimodal}
Mandela Patrick, Yuki~M. Asano, Polina Kuznetsova, Ruth Fong, João~F.
  Henriques, Geoffrey Zweig, and Andrea Vedaldi.
\newblock Multi-modal self-supervision from generalized data transformations.
\newblock \emph{arXiv:2003.04298}, 2020.

\bibitem[Qiu et~al.(2006)Qiu, Lee, Lin, and Yu]{bearingdata}
Hai Qiu, Jay Lee, Jing Lin, and Gang Yu.
\newblock Wavelet filter-based weak signature detection method and its
  application on rolling element bearing prognostics.
\newblock \emph{Journal of Sound and Vibration}, 2006.

\bibitem[Salamon et~al.(2014)Salamon, Jacoby, and Bello]{urban8k}
J.~Salamon, C.~Jacoby, and J.~P. Bello.
\newblock A dataset and taxonomy for urban sound research.
\newblock In \emph{22nd {ACM} International Conference on Multimedia
  (ACM-MM'14)}, 2014.

\bibitem[Tonekaboni et~al.(2021)Tonekaboni, Eytan, and Goldenberg]{tnc}
Sana Tonekaboni, Danny Eytan, and Anna Goldenberg.
\newblock Unsupervised representation learning for time series with temporal
  neighborhood coding.
\newblock In \emph{Proceedings of the International Conference on Learning
  Representations (ICLR)}, 2021.

\bibitem[van~den Oord et~al.(2018)van~den Oord, Li, and Vinyals]{cpc}
A{\"{a}}ron van~den Oord, Yazhe Li, and Oriol Vinyals.
\newblock Representation learning with contrastive predictive coding.
\newblock \emph{arXiv:1807.03748}, 2018.

\bibitem[Wang and van~den Oord(2021)]{audiosimclr}
Luyu Wang and Aaron van~den Oord.
\newblock Multi-format contrastive learning of audio representations.
\newblock \emph{arXiv:2103.06508}, 2021.

\bibitem[{Zhao} et~al.(2020){Zhao}, {Li}, and {Chen}]{imperfect}
S.~{Zhao}, X.~{Li}, and Y.~C. {Chen}.
\newblock A classification framework using imperfectly labeled data for
  manufacturing applications.
\newblock In \emph{2020 25th IEEE International Conference on Emerging
  Technologies and Factory Automation (ETFA)}, 2020.

\end{thebibliography}
\newpage
\appendix
%-------------------------------------------------------------------------
\section{}\label{apd:first}
\subsection{ContrNP regression}\label{apd:a1}
\begin{figure}[ht]
\begin{center}
\includegraphics[width=1\textwidth]{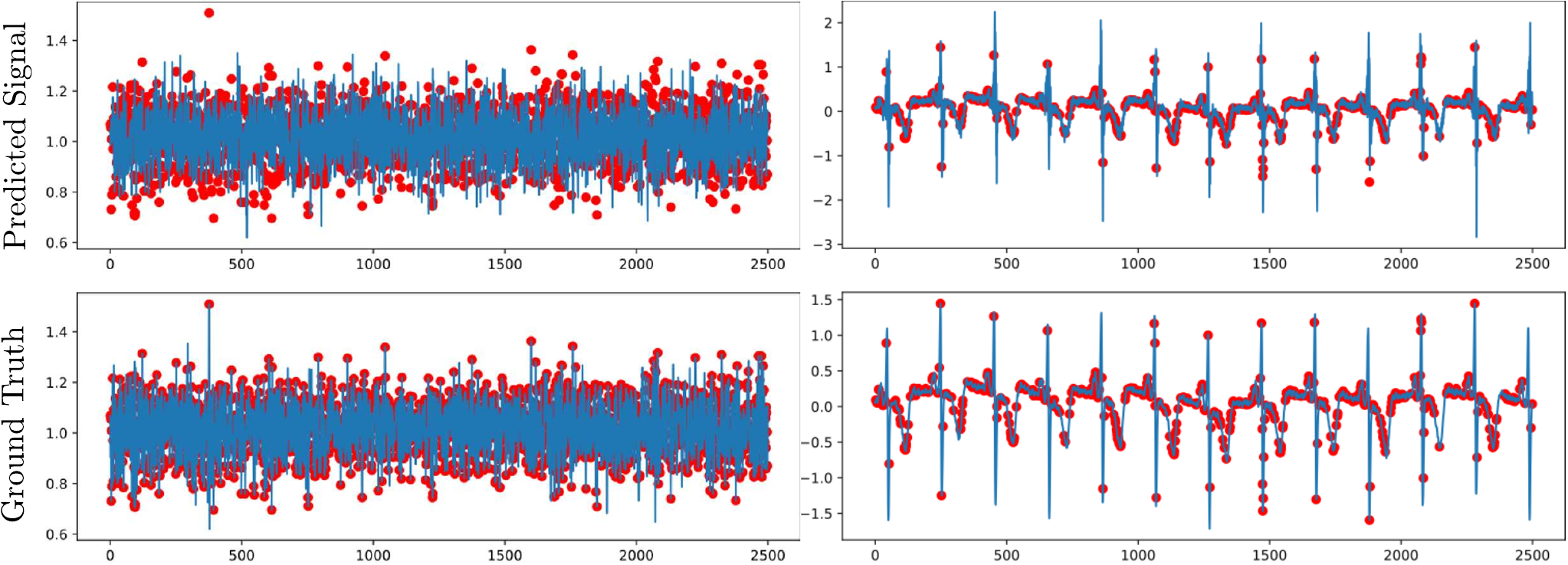}
\caption{ContrNP regression on the IMS (left) and AFDB (right) datasets. The ground truth represents the original signal and the predicted signal is the output of the regression using contrastive neural processes.}
\label{fig:4}
\end{center}
\end{figure}

Contrastive neural processes use a dual learning objective. The contrastive loss ensures the class representations are separable and the Negative Log Likelihood (NLL) loss is used to decrease the within cluster distance (improve clusterability). Figure \ref{fig:4} demonstrates that our method learns successfully to forecast time series. Contrastive neural processes are also able to predict datasets such as the IMS bearing which have a complicated, noisy structure.

\subsection{Detailed Experiment Settings}\label{apd:a2}
Our approach is based on the convolutional neural processes \footnote{\url{http://yanndubs.github.io/Neural-Process-Family}} implementation from \citep{convcnp}. For the AFDB dataset we follow the same pre-processing steps as in \citep{tnc}. We further separate the ECG signals into 5 parts each, normalize and shuffle. In the IMS bearing dataset we only use the third bearing in the first test \citep{imperfect}. The Urban8k dataset is split into 10 individual folds and has 10 classes. For each class we extract a sound sequence of size 100,000 and perform 10-fold cross validation. All of the experiments where conducted with an encoding size of 128 and a window size 2500. SimCLR is implemented with Time Stretch and Frequency Mask augmentations. After experimentation, "Gain" (multiplying the amplitude) and "Shifting" (shift audio backwards/forwards) are able to perform better on the IMS dataset and achieve an accuracy of 69.1\%. On the other hand, our approach does not require any augmentations to achieve an accuracy of 73.6\%. 

\subsection{Selecting the correct hyperparameter $\lambda$}\label{apd:a3}
\begin{figure}[ht]
\begin{center}
\includegraphics[width=1\textwidth]{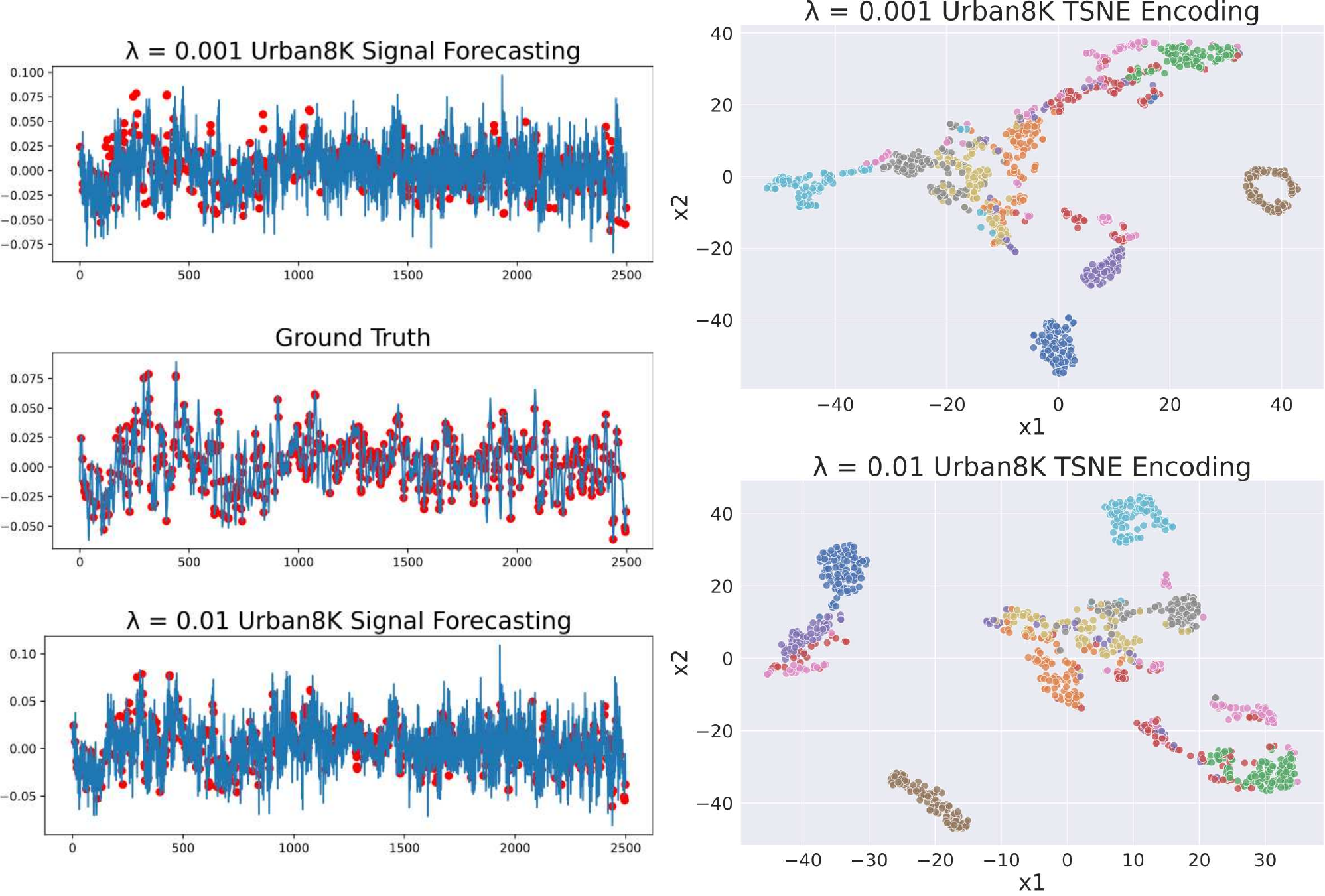}
\caption{ContrNP regression on the Urban8K dataset (left) and 2D TSNE encoding representations (right) for different values of $\lambda$. Experiments were conducted using the first fold as a test set and the remaining for training.}
\label{fig:5}
\end{center}
\end{figure}

The hyperparameter $\lambda$ is the most important parameter in our ContrNP approach. It operates similar to regularization by controlling the objective of the loss function. A higher value will correspond to a stronger forecasting objective and cause our model to operate similar to CPC. On the other hand, a small value for $\lambda$ indicates a more contrastive approach and the model operates closer to SimCLR. The target is to select a value that will benefit from both objectives and ensure a better generalization for the model. Experimentally, a good value is typically 0.01 which emphasizes contrastive learning for faster optimization but also produces a good regression output. Event though a value of 0.01 and 0.001 is a $10\times$ difference, the forecasting in Figure \ref{fig:5} is not greatly affected. On the other hand, the performance improves greatly for higher values of lamdba as seen from the encodings.

%\newpage
\subsection{ContrNP Pseudocode}
\begin{algorithm}[ht]
\begin{lstlisting}
# oc, ot: context and target observation sets
# a, r : sampling range of the context set and representations
# g, d : encoder and decoder parts of any neural process
for e in epochs:
    oc1, oc2 = sample(f([:a])), sample(f([:a])) # m=2 augmentations
    ot1, ot2 = sample(f), sample(f)
    
    rc1,rc2 = encode(oc1),encode(oc2)
    rt1,rt2 = induce(rc1,ot1),induce(rc2,ot2)
    p_y = d(xt,rt)

    loss = sumlogprob(p_y)  + sim(rc1,rc2)/sum(sim(rc,R))
    loss.backward()
    
def encode(o):
    for x,y in o:
        r.append(g(x,y)) # encode each x,y pair
    return mean(r) # aggregate representations
    
\end{lstlisting}
\caption{PyTorch pseudocode of ContrNP for a function $f(t)$ where $m=2$.}
\label{alg:1}
\end{algorithm}
\subsection{Additional Experiments}
\begin{figure}[ht]
\begin{center}
\includegraphics[width=1\textwidth]{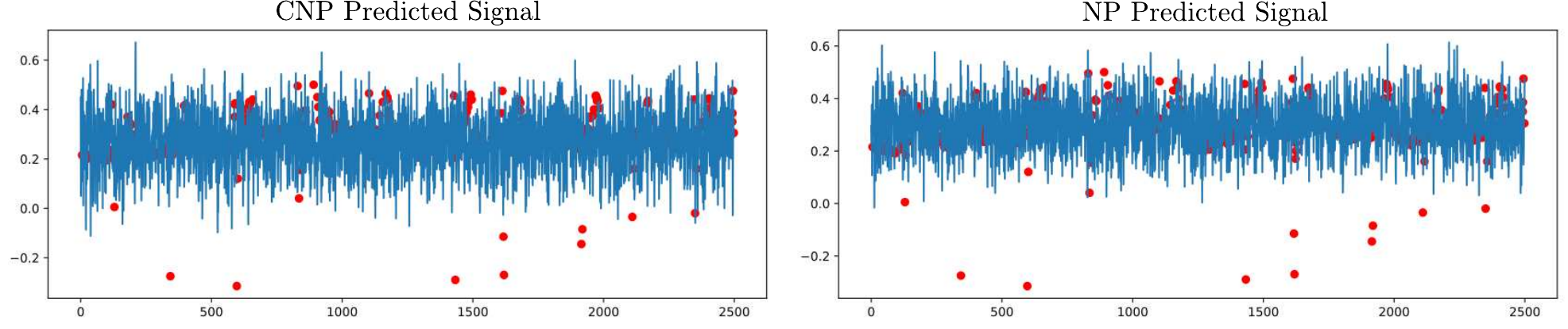}
\caption{Predicted regression signal on AFDB using CNP (left) and NP (right).}
\label{fig:6}
\end{center}
\end{figure}

Using different NPs as a base method, the largest performance drop occurs on the AFDB experiments. The new models achieve 74.7\% using CNP and 72.3\% with NP. This is significantly lower than ConvCNP with a 94.2\% accuracy and can be attributed to the periodicity of the dataset which greatly benefits from the property of translational equivariance. More importantly, even though the regression function performs poorly (Figure \ref{fig:6}), especially considering our original implementation (Figure \ref{fig:4}), the accuracy of our method remains high when compared to the state of the art, due to our dual learning objective. We also perform experiments on images by basing our implementation on the on-grid ConvCNP with full translation equivariance setting \citep{convcnp}. We achieve a CIFAR-10 accuracy of 82.2\% using a ResNet with just over 1M parameters. This shows the flexibility of our approach that is data agnostic and does not need specialized augmentation pipelines.

\end{document}